# AN EMPIRICAL MODEL OF ACKNOWLEDGMENT FOR SPOKEN-LANGUAGE SYSTEMS

David G. Novick and Stephen Sutton
Interactive Systems Group
Department of Computer Science and Engineering
Oregon Graduate Institute
20000 N.W. Walker Rd.
P.O. Box 91000
Portland, OR 97291-1000
novick@cse.ogi.edu

## Abstract

We refine and extend prior views of the description, purposes, and contexts-of-use of acknowledgment acts through empirical examination of the use of acknowledgments in task-based conversation. We distinguish three broad classes of acknowledgments (other→ackn, self→other→ackn, and self+ackn) and present a catalogue of 13 patterns within these classes that account for the specific uses of acknowledgment in the corpus.

## 1 MOTIVATION

This study is motivated by the need for better dialogue models in spoken-language systems (SLSs). Dialogue models contribute directly to the interaction by providing inter-utterance coherence. Fluent understanding and use of acknowledgments should improve spoken-language systems in at least the following ways:

- *Preventing miscommunication.* Acknowledgments are an important device for establishing mutual knowledge and signaling comprehension. Early detection and correction of cases of miscommunication and misunderstanding should prevent failure that would otherwise have been even more catastrophic.

- *Improved naturalness.* Acknowledgments are a prominent feature of human-human dialogue. Supporting the use of acknowledgments for both the system and the user will emphasize the "naturalness" of interfaces and improve their utility.

- *Dialogue control.* Humans cope with dialogue control (e.g., turn-taking) with seemingly little or no effort. Acknowledgments form an intricate relationship with dialogue control mechanisms. Understanding these control mechanisms is central to the development and success of spoken language systems in order to "track" dialogues and determine appropriate system actions.

- *Improved recognition.* To the extent that a dialogue model can narrow the range of possible contexts for interpretation of a user's utterance, a spoken-language system's speech recognition performance will be improved (Young et al., 1989).

We seek to refine and extend prior views of the description, purposes, and contexts-of-use of acknowledgment acts through empirical examination of the use of acknowledgments in task-based conversation. In particular, we seek to describe systematically (1) the communicative value of an acknowledgment and (2) the circumstances of its use. The scope of our inquiry involves spoken interaction. We present a catalogue of types of acknowledgment. This catalogue is based on a process model of acknowledgment that explains instances of these acts in a corpus of task-based conversations.

## 2 RELATED WORK

Clark and Schaefer (1989) suggested that acknowledgments are an important component of a larger framework through which communicating parties provide evidence of understanding. Conversants have a range of means, which vary with respect to strength, for indicating acceptance of a presentation. These include continued attention, initiation of the next relevant contribution, acknowledgment, demonstration. and display.

Thus acknowledgments are common linguistic devices used to provide feedback. Broadly speaking, acknowledgments are responsive acts.[1] That is, they are usually uttered in (possibly partial) response to a production by another speaker; acknowledgment acts express beliefs and intentions of one conversant with respect to the mutuality of prior exchanges involving some other conversant. The intended perlocutionary effect of an acknowledgment act is generally the perception of mutuality of some belief.

---

[1] A notable exception is the self-acknowledgment which will be discussed shortly

In previous research, the function of acknowledgments has been most readily characterized in terms of attention, understanding and acceptance on the recipient's behalf (Kendon, 1967; Schegloff, 1982). In addition, it has been suggested that they serve to facilitate active participation in dialogues and promote "smooth" conversations (Duncan and Fiske, 1987).

Schegloff (1982) described two main types of acknowledgment: continuers and assessments. Continuers, such as "uh huh," "quite," and "I see," act as bridges between units. Conversants use acknowledgments as continuers to signal continued attention and to display the recipient's understanding that the speaker is in an extended turn that is not yet complete. Moreover, continuers indicate the turning down of an opportunity to undertake a repair subdialogue regarding the previous utterance or to initiate a new turn. Assessments—such as "oh wow" and "gosh, really?"— are essentially an elaboration on continuers. That is, they occur in much the same environment and have similar properties to continuers, but in addition express a brief assessment of the previous utterance.

Empirical analysis of conversations has indicated that the occurrence of acknowledgments is not arbitrary. Acknowledgments mostly occur at or near major grammatical boundaries, which serve as transition-relevance places for turn-taking (Sacks et al., 1974; Hopper, 1992). In particular, work by Orestrom (1983) and Goodwin (1986) suggested a tendency for continuers to overlap with the primary speaker's contribution, in such a way that they serve as bridges between two turn-constructional units. Assessments, on the other hand, are generally engineered without overlap. Goodwin suggested that conversants make special efforts to prevent assessments from intruding into subsequent units. That is, the speaker typically delays production of the subsequent unit until the recipient's assessment has been brought to completion.

Clearly, acknowledgments are an important device for providing evidence of understanding and for avoiding miscommunication between parties. Just as next-relevant-contributions include the entire range of potential task or domain actions, the task-based role of acknowledgments can be differentiated within their class as acceptances. Beyond continuers and assessments, we will demonstrate that acknowledgments incorporate a larger set of conversational actions, many of which relate to coherence of multi-utterance contributions.

## 3 DIALOGUE ANALYSIS

In this section, we describe the task characteristics and the corpus used for this study, present a theoretical model of acknowledgment acts in task-based dialogue, and present an analysis of acknowledgment acts based on corpus material.

### 3.1 THE VEHICLE NAVIGATION SYSTEM CORPUS

The corpus we analyzed was collected by U S WEST Advanced Technologies in the domain of a vehicle navigation system (VNS). A VNS is intended to provide travel directions to motorists by cellular telephone: the system interacts with the caller to determine the caller's identity, current location and destination, and then gives driving directions a step at a time under the caller's control. U S WEST collected the dialogues through a Wizard-of-Oz experiment (Brunner et al., 1992) in which the wizard was free to engage in linguistically unconstrained interaction in the VNS task. Each of the 21 subjects performed three tasks in the VNS domain. As a whole, the corpus contained 2499 turns and 1107 acknowledgments.

### 3.2 A TASK-BASED MODEL OF ACKNOWLEDGMENT ACTS

The generally accepted view of acknowledgments, as noted earlier, distinguishes between two classes—namely continuers and assessments (Schegloff, 1982). Indeed, there were many occurrences of continuers (and a few assessments) in the VNS corpus. However, our analysis suggests that acknowledgments perform functions beyond these two classes. For instance, we observed several instances of acknowledgment being used at the beginning of a turn by the same speaker. This contrasts with the notions of continuers and assessments which, by definition, occur as unitary productions in the context of extended turns by another speaker. Clearly, an acknowledgment occurring at the beginning of a turn is not serving as a prompt for the other speaker to continue.

To account for the evidence provided by the VNS corpus, we propose to extend Schegloff's classification scheme into a task-based model of acknowledgment acts. This model formalizes the meaning and usage characteristics of acknowledgments, based on an analysis of what acknowledgments mean and when acknowledgments are used in the VNS dialogues.

A useful way of looking at the role of acknowledgments in the context of turns is to consider the basic structural context of exchanges. We begin by reviewing the concept of an adjacency pair (Schegloff and Sacks, 1973; Clark and Schae-fer, 1989). An adjacency pair is formed by two consecutive utterances that have a canonical relationship, such as question-answer and greeting-greeting. An acknowledgment can be produced as the second phase of an adjacency pair or following a complete adjacency pair; in each case, the utterance may contain multiple acceptances. Of course, an acknowledgment can be produced also as a single turn that does not relate to an adjacency pair. Thus, based on exchange structure one can distinguish three broad structural classes of acknowledgments:[2]

- *Other→ackn*, where the acknowledgment forms the second phase of an adjacency pair;
- *Self→other→ackn*, where Self initiates an exchange, Other (eventually) completes the exchange, and Self then utters an acknowledgment; and
- *Self+ackn*, where Self includes an acknowledgment in an utterance outside of an adjacency pair.

In the *other→ackn* class, the exchange is a basic adjacency pair; Other's act will be composed of a single turn. In the *self→other→ackn* class, the exchange initiated and eventually acknowledged by Self may be composed of multiple turns, with multiple utterances by both Self and Other. In the *self+ackn* class, the acknowledgment occurs in a single, extended turn by Self that may contain multiple utterances.

## 3.3 A CATALOGUE OF ACKNOWLEDGMENT ACTS IN TASK-BASED DIALOGUE

In this section, we elaborate the structural classes of acknowledgment through a catalogue of patterns of speech acts that occur in each class. This catalogue provides broad coverage of patterns typically encountered in task-oriented discourse. These patterns describe the context of acknowledgments in terms of exchanges and are derived from utterances in the VNS corpus. Each act in an exchange is represented in terms of speech-act verbs based on the set described by Wierzbicka (1987). Table 1 summarizes the speech-act patterns in the catalogue. In the following sections, we will consider each of the structural classes in turn and provide examples of selected patterns from the VNS corpus. We also consider embedded exchanges, in which basic patterns are used to build more complex patterns.

**3.3.1 Other→Ackn** Acknowledgments in the *other→ackn* class relate principally to the immediately antecedent utterance as opposed to the prior exchange, which is covered by the *self→other→ackn* class. In Clark and Schaefer's (1989) terms, Self's acknowledgment in the *other→ackn* class serves as the acceptance phase for Other's presentation. As listed in Table 1, the canonical *other→ackn* patterns are *inform→ackn*, *inform→ackn+mrequest*, *request→ackn+inform*, *mdirect→ackn* and *preclose→ackn*.[3] In each of these cases, the first turn is by Other and the second turn is Self's acknowledgment. In some cases, Self's turn also extends to include other significant utterances. We illustrate the *other→ackn* class through examination of the *inform→ackn* and *inform→ackn+mrequest* patterns.

### Inform→Ackn

The *inform→ackn* pattern covers cases where Other performs an *inform* act and Self responds with an acknowledgment of that act. In the following example[4] of an *inform→ackn*, the wizard gives directions to the user, who acknowledges these directions. This is an example of an acknowledgment that Schegloff (1982) would call a continuer.

### Example 1 (U6.3.1)[5]

(1.1) Wizard: On Evans, you need to turn left and head West for approximately three quarters of a mile to Clermont.
(1.2) User: **Okay**.
(1.3) Wizard: And, um, on Clermont you turn left, heading South for about two blocks to Iliff.

Here, the "okay" at turn 1.2 indicates the user's acceptance of the wizard's utterance. That is, the acknowledgment implies that the user understood information given by the wizard— more emphatically than a simple next-relevant-contribution response. Use of the acknowledg-ment would be strong evidence of understanding in Clark and Schaefer's (1989) terms. An important point to stress here is that the wizard cannot rely on the user necessarily having received the information that was actually conveyed or formed the intended interpretation. Rather, the wizard is left with the user's response indicating that the user was apparently satisfied with the wizard's original presentation.

---

[2]The notation for structural class names indicates turns delimited by arrows (→). Acts combined within a turn are joined with a plus (+) symbol. *Other* and *self* refer to non-acknowledgment acts by the respective conversants. "Self" refers to the party producing the acknowledgment; "Other" is the other party.

[3]The *mrequest* and *mdirect* acts are forms of meta-act in which the speaker initiates a clarification subdialogue or otherwise explicitly addresses the control of the conversation; *mrequest* and *mdirect* are extensions of Wierzbicka's (1987) speech-act categories following Novick's (1988) meta-act analysis.

[4]In the examples, the acknowledgment of principal interest is highlighted.

[5]All examples are extracted from a corpus of telephone dialogues from a task-oriented "Wizard-of-Oz" protocol collection study described in Section 3.1. The examples in this paper are notated with the corpus dialogue reference number and each turn is numbered for purposes of reference.

| $Other{\rightarrow}Ackn$ | $Self{\rightarrow}Other{\rightarrow}Ackn$ | $Self{+}Ackn$ |
|---|---|---|
| inform→ackn | inform→ackn→ackn | inform+ackn+inform |
| inform→ackn+mrequest | request→inform→ackn | mrequest+ackn |
| request→ackn+inform | mrequest→inform→ackn | mdirect+ackn |
| mdirect→ackn | mdirect→ackn→ackn | |
| preclose→ackn | | |

Table 1: A Summary of Speech-Act Patterns for Structural Classes of Acknowledgment

**Inform→Ackn+MRequest**

The *inform→ackn+mrequest* class represents a significant functional variation on the *inform→ackn* class just considered. It covers cases where Other performs an inform act, Self responds with an acknowledgment of that act and then goes on to seek clarification of the content of the inform act. Requests for clarification are kinds of meta-act because they are concerned with aspects of dialogue control rather than the task itself. That is, requests for clarification are concerned with the specifics of old information rather than seeking to elicit largely new information—unlike request-inform acts.

**Example 2 (U4.3.1)**

(2.1) Wizard: Okay. Then you want to go north on Speer Boulevard for one and one half miles to Alcott Street.

(2.1) User: **Okay**. I want to go right on Speer?

(2.2) Wizard: It will be a left.

In this example, the repair is a potential request for specification (Lloyd, 1992). That is, the user's clarification at 2.2 ("I want to go right on Speer?") focuses on information which was missing from the surface structure of the original inform act but which is potentially available—namely "right" instead of "north."

**3.3.2 Self→Other→Ackn** Acknowledgments in the *self→other→ackn* class relate to the previous exchange, rather than just the previous utterance. Broadly speaking, they express the current state of the dialogue in addition to embodying the functionality of *other→ackn* acknowledgments. That is, they explicitly mark the completion of the antecedent exchange and indicate that the dialogue will either enter a new exchange or resume an existing exchange. Furthermore, *self→other→ackn* acknowledgments signal understanding and acceptance of both the previous exchange and the previous utterance. The canonical patterns of the *self→other→ackn* class, as listed in Table 1, include *inform→ackn→ackn*, *request→inform→ackn*, *mrequest→inform→ackn* and *mdirect→ackn→ackn*. We illustrate the *self→other→ackn* class through examination of the *request→inform→ackn* pattern.

**Request→Inform→Ackn**

The *request→inform→ackn* class covers cases where Self requests an *inform* act of Other. Other then performs that *inform* act and Self acknowledges. Note that the acknowledgment in this case follows a completed request-inform adjacency pair. Earlier, we mentioned that question-answer adjacency pairs can be regarded as special cases of request-inform adjacency pairs (Searle, 1969). In the following example, the wizard requests the user's start location. The user satisfies this request by communicating the desired information and the wizard then acknowledges. Here the acknowledgment at 3.3 serves to indicate acceptance (that is, receipt, understanding and agreement) of the user's inform act and is a signal that the request initiated by the wizard at 3.1 has been satisfied and thus the exchange is complete.

**Example 3 (U2.1.1)**

(3.1) Wizard: Okay and uh, what's your starting location?

(3.2) User: I'm at 36th and Sheridan at the Park-n-Ride.

(3.3) Wizard: **Okay**, one moment please.

**3.3.3 Self+Ackn** Self-acknowledgments occur when Self issues an acknowledgment following some action (either verbal or physical) performed by Self. These are not responsive acts, unlike other acknowledgment usages considered; however, they are still closely tied with the idea of establishing mutual beliefs. The canonical patterns, as listed in Table 1, include *inform+ackn+inform*, *mrequest+ackn*, and *mdirect+ackn*. We illustrate the *self+ackn* class through examination of the *inform+ackn+inform* pattern.

### Inform+Ackn+Inform

In this pattern, Self uses an acknowledgment in the middle of an extended turn. Consider the following example:

### Example 4 (U5.3.1)

(4.1) Wizard: All right, um, the first thing you need to do is go South on Logan Street for one and a half miles to Evans Avenue. Then turn left on Evans Avenue and go one and a quarter miles to South Josephine Street. **Okay**, then you'll turn left on South Josephine Street. Nineteen Forty South Josephine is within the first block.

This particular self-acknowledgment is very similar to a continuer—indeed it may be regarded as a *self-continuer*. The wizard's acknowledgment in this example represents a sort of temporizing, a chance for the wizard to "catch his mental breath." For the user, this sort of "Okay" thus signals that the wizard intends to continue his turn. This is functionally distinct from a meta-request of the form "Okay?" because there is no rising intonation and the wizard does not wait for a response. In fact, use of a self-acknowledgment at the end of a turn would be peculiar.

**3.3.4 Embedded Exchanges** We noted earlier that basic patterns can used to build more complex patterns. This can lead potentially to variations in patterns of acknowledgments. In particular, it is possible to observe cascades of acknowledgments as nested exchanges are "popped" one by one. Simple acts may be replaced by more complex exchanges, so that an inform act may be replaced by an exchange that accomplishes an inform via a sequence of informs, clarifications and acknowledgments. In this section we will consider one of the variations encountered in the VNS corpus; where an *inform→ackn→ackn* replaces the *inform* act in an *inform→ackn* sequence. In the following example, there are three successive acknowledgment acts. The first acknowledgment at 5.2 is accompanied by a verbatim response by the user. It is the second phase of the *inform→ackn* adjacency pair, indicating understanding and acceptance of the wizard's *inform* act in which a direction was clarified. The second acknowledgment, issued by the wizard at 5.3, marks the completion of the *inform→ackn* exchange. That is, the wizard recognizes that it is his or her turn yet has nothing more to add, so indicates passing up the turn with an acknowledgment. The third acknowledgment, issued by the user at 5.4, is associated with the user recognizing that the wizard has finished clarifying directions; the user thus acknowledges this embedded *inform* act. The user then indicates satisfaction and approval of the wizard's directions with the assessment "Sounds good."

### Example 5 (U6.2.1)

(5.1) Wizard: Okay, it was, um, on Evans it's three and three quarter miles to Jasmine.
(5.2) User: Three, **okay**.
(5.3) Wizard: **Okay**.
(5.4) User: **All right**, sounds good.

## 4 CONCLUSION

Why is a conversation-analytic study of acknowledgment useful in the development of spoken language systems? SLSs developers face the dual challenges of creating both domain-based dialogues and repair-oriented dialogues. Lacking systematic mechanisms for natural maintenance of mutuality, SLSs tend to rely on domain structures—producing rather stolid interaction. The most advanced systems incorporate repair acts, but are unable to relate the repairs to the main dialogue structures in a natural way. The acknowledgment model described in this paper provides a systematic method of maintaining mutuality of knowledge for both domain and control information.

More concretely, using this model SLSs can account for acknowledgments by both user and system. The corpus evidence suggests that users' utterances in unconstrained dialogues contain many instances of acknowledgment. In interpreting these utterances, identification of the appropriate acknowledgment function affects the state of the dialogue model and thus plays an important role in determining an appropriate response by the system. In producing such responses, the acknowledgment model can provide structurally appropriate utterances. The fundamental idea is to produce contextually appropriate acknowledgments that advances the dialogue seamlessly with respect to both domain and control functions. That is, the system needs to give the right signals at the right time.

The evidence of the U S WEST VNS corpus suggests that understanding and production of domain and control utterances are closely linked; they thus cannot be implemented as independent mechanisms in an SLS. For example, giving directions involves presenting large amounts of information for which an installment approach often proved effective. Typically the user was given the opportunity to choose the style of presentation of directions, either step-by-step or all at once. The choice of presentation method by the conversants was a dynamic one: in cases where it became apparent that the user was experiencing difficulties with either hearing or understanding directions, the wizard often resorted to the step-by-step approach. This form of repair changed the process of interaction so that the comprehension of each installment was verified before proceeding with the next.

The conversants in the VNS corpus displayed relatively higher rates of use of acknowledgment in repair situations or when unplanned events arose (e.g., the user had gotten lost). Intuitively, people make more effort to establish mutual knowledge when it is apparent that miscommunication has occurred than at other times; their certainty criterion for mutuality (Clark and Marshall, 1981) is raised as a result of the need for repair. This suggests that a facility for acknowledgment is an important element in the development of robust SLSs because use of acknowledgment is most critical precisely when the conversation has gone awry.

We are currently developing a computational model of acknowledgment based on the empirical work presented in this paper. This model is intended for integration into a SLS where it will serve both to predict when acknowledgments are appropriate from the system and when to expect acknowledgments from the user. Briefly, determining the applicability of an acknowledgment involves interpreting exchanges in terms of speech acts and then mapping these speech-act patterns onto the acknowledgment classes described. This, we believe, will facilitate improved SLS robustness through achievement of a greater degree of mutual understanding and provide a more natural and intuitive interaction. The utility and implementation of the empirical model will be the focus of a later paper.

## 5 ACKNOWLEDGMENTS

This work was supported by U S WEST Advanced Technologies and the Oregon Advanced Computing Institute.